\documentclass[conference,letterpaper]{IEEEtran}
\IEEEoverridecommandlockouts
\usepackage{cite}
\usepackage{amsmath,amssymb,amsfonts}
\usepackage{algorithmic}
\usepackage{graphicx}
\usepackage{textcomp}
\usepackage{xcolor}
\usepackage{url}
\usepackage{pgfplots}
\usepackage{filecontents}

\usepackage{float} 

\usepackage{geometry}
\begin{document}

\newgeometry{top=54pt, left=54pt, right=54pt, bottom=54pt}

\title{Exploring Attention-Driven Deep Learning for Accurate Lung Nodule Segmentation}

\author{
\IEEEauthorblockN{Nima Shafiei Rezvani Nezhad\IEEEauthorrefmark{1}, Meysam 
Mansouri\IEEEauthorrefmark{2}, Saeid Ghayour \IEEEauthorrefmark{3}, Ruhollah Abolhasani\IEEEauthorrefmark{4}, Shahla Azizi\IEEEauthorrefmark{5}\thanks{*Corresponding author: Shahla.Alikamar@emu.edu.tr}}

\IEEEauthorblockA{\IEEEauthorrefmark{1}Department of Computer Science and Engineering - DISI, University of Bologna, Bologna, Italy \\
Email: nima.shafieirezvani@studio.unibo.it}

\IEEEauthorblockA{\IEEEauthorrefmark{2}Department of Electronic Engineering, University of Rome Tor Vergata, Rome, Italy \\
Email: mansouri.meysam@gmail.com}

\IEEEauthorblockA{\IEEEauthorrefmark{3}Department of Computer Science, Virginia, USA \\
Email: saeid24@vt.edu}

\IEEEauthorblockA{\IEEEauthorrefmark{4}Department of Applied Health Sciences, Brock University, Ontario, Canada \\
Email: rabolhasani@brocku.ca}
\IEEEauthorblockA{\IEEEauthorrefmark{5}Department of Electrical and Electronics Engineering, Eastern Mediterranean University, Gazimağusa, Mersin 10, TURKEY \\
Email: Shahla.Alikamar@emu.edu.tr}

}

\maketitle

\noindent\begin{minipage}{\columnwidth}
\scriptsize
\textbf{IEEE publication notice.} Published as N. Shafiei Rezvani Nezhad,
M. Mansouri, S. Ghayour, R. Abolhasani, and S. Azizi,
``Exploring Attention-Driven Deep Learning for Accurate Lung Nodule
Segmentation,'' in \emph{2025 47th Annual International Conference of the
IEEE Engineering in Medicine and Biology Society (EMBC)}, 2025.
DOI: \url{https://doi.org/10.1109/EMBC58623.2025.11254801}.

\smallskip
\textcopyright{} 2025 IEEE. Personal use of this material is permitted.
Permission from IEEE must be obtained for all other uses, in any current
or future media, including reprinting/republishing this material for
advertising or promotional purposes, creating new collective works, for
resale or redistribution to servers or lists, or reuse of any copyrighted
component of this work in other works.
\end{minipage}
\par\vspace{0.7em}

\setlength{\topmargin}{-0.55in}
\setlength{\textheight}{9.25in}
\setlength{\oddsidemargin}{-0.25in}
\setlength{\evensidemargin}{-0.25in}

\begin{abstract}
This study explores the application of various deep learning models for the segmentation of lung nodules using LIDC-IDRI. Unlike traditional approaches that utilize the full dataset, our work emphasizes the efficacy of training models on a filtered subset of 356 samples. Novel configurations, including attention mechanisms and advanced preprocessing strategies, were employed to optimize segmentation accuracy. Among the models evaluated, the DPLinkNet50 with a Channel Attention Bridge and ResNet backbone demonstrated the highest performance with a Dice score of 0.86, significantly outperforming conventional architectures. This work underscores the potential of leveraging data efficiency and tailored architectures in achieving robust segmentation performance, paving the way for improved computer-aided diagnosis in clinical settings.
\end{abstract}


\section{Introduction}
\label{submission}

Lung cancer remains the leading cause of cancer-related mortality worldwide, accounting for nearly 25\% of all cancer deaths \cite{bianconi2021}. Early detection and diagnosis are critical for improving survival rates, as the prognosis depends significantly on the stage of the disease at diagnosis \cite{farag2013}. Pulmonary nodules, which often manifest as solid, sub-solid, or ground-glass opacities on CT scans, serve as early indicators of lung cancer. However, their diverse morphologies, intensities, and adjacency to complex anatomical structures pose significant challenges for accurate detection and segmentation \cite{keetha2020}.

Computed tomography (CT) is the gold standard imaging modality for lung nodule detection, offering high sensitivity and resolution. Nevertheless, manual segmentation of nodules by radiologists is time-consuming, labor-intensive \cite{bianconi2021, zheng2018}. This has driven the adoption of automated and semi-automated approaches, with deep learning playing a pivotal role in advancing segmentation accuracy and efficiency.

Deep learning, particularly convolutional neural networks (CNNs), has become a cornerstone in medical image segmentation. Architectures like U-Net and its variants have achieved notable success in delineating pulmonary nodules due to their encoder-decoder structure and ability to capture fine-grained details \cite{ronneberger2015, suji2023}. Moreover, the integration of attention mechanisms, such as in Channel Attention Bridges, has further improved segmentation performance by focusing on the most relevant features in the input images \cite{keetha2020, zhao2019}.

Recent advancements have introduced innovations like multi-task learning frameworks (e.g., PN-SAMP) that combine segmentation, attribute extraction, and malignancy prediction within a unified model, enhancing interpretability and diagnostic value \cite{wu2020}. Other approaches, such as pyramid deconvolutional neural networks, have addressed the challenge of segmenting nodules with complex textures and boundaries by fusing multi-scale information \cite{zhao2019}. Furthermore, methods based on reinforcement learning, \cite{tian2020}, have explored iterative refinements to segmentation outputs, mimicking radiologists' multi-step annotation processes.

Despite these advancements, segmentation accuracy remains constrained by the variability and noise inherent in medical imaging datasets. Traditional methods like level sets and active contour models struggle to handle this variability effectively \cite{farag2013, zheng2018}. Deep learning models address these limitations, but their performance often depends on the quality and quantity of training data. In this context, transfer learning with pre-trained models has emerged as a powerful tool to optimize network performance and reduce computational requirements \cite{suji2023}.

\subsection{Related Works}
\label{related-works}

Recent studies have introduced innovative methods that significantly enhance segmentation accuracy and robustness. A comparison of these methods, including their methodologies, datasets, strengths, and limitations, is presented in Table~\ref{tab:comparison-methods}.

Miao et al. \cite{miao2025} proposed a deep learning-based nomogram that integrates intrathoracic fat (ITF) and pulmonary nodule features to classify malignant nodules. By combining imaging biomarkers and predictive models, the approach achieved high diagnostic precision. However, its reliance on ITF features limits its generalizability to cases without distinct fat delineation.

Shi et al. \cite{shi2024} introduced MAST-UNet, which combines Local Aware Attention and Pixel Transformer modules for enhanced semantic feature extraction. This architecture demonstrated robust segmentation performance, particularly on small nodules, when applied to the LIDC-IDRI dataset. The complexity of the architecture, however, increases training time and requires a large dataset.

Zhang et al. \cite{zhang2025} developed SPC-UNet, incorporating pyramidal squeeze attention for multi-scale contextual information. This model achieved notable accuracy in handling nodules with complex textures and boundaries. Nonetheless, the computational requirements of SPC-UNet may restrict its deployment in real-time clinical applications.

Rajasekar et al. \cite{rajasekar2025} proposed GAME, a generative autoencoder with attention mechanisms for unsupervised lung nodule segmentation. GAME is highly effective in handling low-quality images and limited datasets. Despite these advantages, the reliance on unsupervised learning may result in lower accuracy compared to supervised methods.

Wang et al. \cite{wang2023} presented CAUNet, an architecture that employs channel attention mechanisms to improve feature selection and reduce false positives in nodule segmentation. CAUNet performs well with non-solid nodules but is sensitive to hyperparameter tuning.

\begin{table}[htbp]
\caption{Comparison of Recent Lung Nodule Segmentation Methods}
\begin{center}
\begin{tabular}{|p{0.05\textwidth}|p{0.10\textwidth}|p{0.05\textwidth}|p{0.08\textwidth}|p{0.10\textwidth}|}
\hline
\textbf{Paper} & \textbf{Methodology} & \textbf{Dataset} & \textbf{Strengths} & \textbf{Limitations} \\
\hline
Miao et al. \cite{miao2025} & DL-based nomogram combining ITF and nodules & LIDC-IDRI & High malignancy classification accuracy & Limited generalizability \\
\hline
Shi et al. \cite{shi2024} & MAST-UNet with Local Attention and Pixel Transformer & LIDC-IDRI & Robust for small nodules & High training cost, needs large data \\
\hline
Zhang et al. \cite{zhang2025} & SPC-UNet with pyramidal squeeze attention & LIDC-IDRI & Effective multi-scale feature extraction & Computationally expensive \\
\hline
Rajasekar et al. \cite{rajasekar2025} & GAME (Generative Autoencoder + Attention) & Custom dataset & Works with low-quality and small datasets & Lower accuracy vs. supervised methods \\
\hline
Wang et al. \cite{wang2023} & CAUNet with channel attention & LIDC-IDRI & Handles non-solid nodules effectively & Sensitive to hyperparameter tuning \\
\hline

\end{tabular}
\label{tab:comparison-methods}
\end{center}
\end{table}

In this study, we present an evaluation of state-of-the-art deep learning models for lung nodule segmentation on the LIDC-IDRI dataset. Our approach leverages architectural innovations like attention mechanisms and channel bridges to achieve robust segmentation with reduced data. 

\section{Proposed Methodology}
\label{proposed-methodology}

\subsection{LIDC-IDRI Dataset}

The LIDC-IDRI dataset \cite{armato2015} is a publicly available collection of 1,018 thoracic CT scans curated to support research in lung nodule detection and diagnosis. Each scan is accompanied by annotations from up to four experienced radiologists, conducted in a two-phase reading process—first independently (blinded), then with access to each other’s findings (unblinded). The dataset includes detailed XML files describing nodules 3 mm or larger, along with attributes such as size, shape, texture, and an estimated malignancy score.

\subsection{Preprocessing Framework}

\begin{figure}[htbp]
    \centering
    \includegraphics[width=\linewidth]{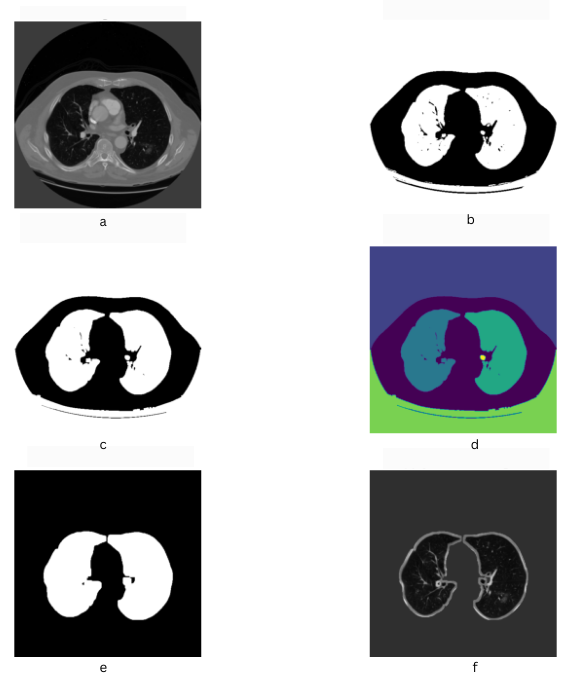} 
    \caption{Illustration of preprocessing steps for lung region segmentation. From A to F: (a) Original CT scan slice, (b) Initial thresholding to separate lung regions from background, (c) Morphological operations (erosion and dilation) to refine the binary mask, (d) K-means clustering for region labeling, (e) Final refined binary lung mask, (f) Application of the final mask to the original scan to extract the lung region of interest (ROI).}
    \label{fig:preprocessing}
\end{figure}

The preprocessing steps are illustrated in Figure~\ref{fig:preprocessing}. These steps ensure the segmentation models receive high-quality input data by reducing noise and extracting relevant lung regions.

\paragraph{Lung ROI Extraction}
To isolate the lung regions, the following steps were employed:
\begin{itemize}
    \item \textbf{K-means Clustering:} Lung tissues were segmented from the background using K-means clustering with \( k = 2 \). The clustering minimizes intra-cluster variance, defined as:
    \[
    J = \sum_{i=1}^{k} \sum_{x \in S_i} \|x - \mu_i\|^2
    \]
    where \( S_i \) represents the \( i \)-th cluster, \( x \) is a data point, and \( \mu_i \) is the centroid of \( S_i \).

    \item \textbf{Morphological Operations:} To refine the lung masks, erosion and dilation were applied. The dilation operation is given by:
    \[
    f'(x, y) = \bigvee_{(s, t) \in B} f(x + s, y + t)
    \]
    where \( B \) is the structuring element.
\end{itemize}

\paragraph{Noise Reduction}
Noise in the CT images was reduced while preserving edges through:
\begin{itemize}
    \item \textbf{Median Filtering:} Applied to smooth the image and reduce impulsive noise.
    \item \textbf{Anisotropic Diffusion:} Modeled as:
    \[
    \frac{\partial I}{\partial t} = \nabla \cdot \left(c(x, y, t) \nabla I \right)
    \]
    where \( c(x, y, t) \) is the diffusion coefficient controlling the smoothing.
\end{itemize}

\paragraph{Normalization and Cropping}
To standardize input data, the following operations were applied:
\begin{itemize}
    \item \textbf{Normalization:} Pixel intensities were scaled to the range [0, 1]:
    \[
    I' = \frac{I - I_{\text{min}}}{I_{\text{max}} - I_{\text{min}}}
    \]
    where \( I_{\text{min}} \) and \( I_{\text{max}} \) represent the minimum and maximum intensity values.
    \item \textbf{Cropping:} Lung regions were extracted using bounding boxes derived from the refined lung masks.
\end{itemize}

This preprocessing pipeline ensures the input images are optimized for segmentation by enhancing relevant features and minimizing noise~\cite{ali2021}.

\subsection{Segmentation Models}

Accurate segmentation of lung nodules in CT images is critical for early diagnosis and treatment of lung cancer. Autoencoder-based architectures have emerged as the cornerstone of deep learning for segmentation tasks, owing to their ability to encode and decode complex spatial patterns in medical images. These models follow an encoder-decoder structure, where:

\begin{itemize}
    \item \textbf{Encoder:} Extracts hierarchical features and compresses them into a latent representation, capturing high-level contextual information.
    \item \textbf{Decoder:} Reconstructs the segmentation map from the latent representation, ensuring fine-grained alignment with the input image.
\end{itemize}

The encoder-decoder framework is often augmented with advanced mechanisms such as skip connections, attention modules, and dilated convolutions to address the unique challenges of lung nodule segmentation, including small nodule detection and complex anatomical boundaries.

\paragraph{DPLinkNet50 + Channel Attention Bridge (CAB)}
DPLinkNet50 integrates ResNet-50 as the encoder with LinkNet’s lightweight decoder. The addition of a Channel Attention Bridge (CAB) enhances feature selection by dynamically weighting feature maps based on their importance:
\[
\mathbf{F}' = \sigma\left(W_2 \cdot \text{ReLU}(W_1 \cdot \text{GAP}(\mathbf{F}))\right) \cdot \mathbf{F}
\]
where \(\mathbf{F}\) represents feature maps, \(\text{GAP}(\cdot)\) is global average pooling, \(W_1\) and \(W_2\) are learnable weights, and \(\sigma\) is the sigmoid activation. The CAB module allows the model to emphasize critical regions, making it highly effective for lung nodule segmentation.

\paragraph{LinkNet}
LinkNet employs a compact encoder-decoder architecture, making it computationally efficient while maintaining strong segmentation performance. Skip connections between the encoder and decoder ensure accurate spatial information transfer, enabling the precise delineation of lung nodule boundaries. \cite{chaurasia2018}

\paragraph{DPLink}
An extension of LinkNet, DPLink incorporates densely connected layers to improve feature reuse and gradient flow. These dense connections enhance the model’s ability to capture complex contextual information, particularly beneficial for segmenting irregularly shaped nodules. \cite{xiong2021}

\paragraph{U-Net Variants}
The U-Net architecture, known for its encoder-decoder design with skip connections, was further enhanced in this study with the following variants:
\begin{itemize}
    \item \textbf{U-Net + Channel Attention Bridge:} Adds a CAB module to focus on relevant feature maps, reducing false positives and improving segmentation accuracy.
    \item \textbf{U-Net + VPM + Channel Attention Bridge:} Combines Visual Positioning Modules (VPM) with CAB to enhance spatial localization and feature selection, enabling better segmentation of small and complex nodules.
\end{itemize}

\paragraph{ResDense with Channel Attention}
This model combines the densely connected architecture of DenseNet with channel attention mechanisms. Dense connections improve feature reuse and gradient propagation, while attention mechanisms enhance the network's focus on critical features, particularly useful for detecting subtle nodule boundaries. \cite{guo2021}

\paragraph{Dilated U-Net}
The dilated U-Net incorporates dilated convolutions in the encoder, which expand the receptive field without increasing the number of parameters. This allows the model to capture multi-scale contextual information, improving its ability to segment nodules adjacent to complex anatomical structures. \cite{xiong2021}

\paragraph{Other Architectures}
In addition to the above, the study also evaluated:
\begin{itemize}
    \item \textbf{EfficientUNet:} A computationally lightweight variant designed for resource-constrained environments, though its accuracy was limited. \cite{baheti2020}
    \item \textbf{AAU-Net:} Incorporates adaptive attention units for feature selection but struggled with small nodule segmentation in our experiments.\cite{chen2022}
    \item \textbf{U-Net3Plus:} Extends U-Net with additional skip connections and dense layers to capture more contextual information.\cite{huang2020}
\end{itemize}

\subsection{Training and Evaluation}

To assess the performance of the implemented segmentation models, rigorous training and evaluation protocols were employed. This section outlines the dataset, training process, and metrics used to quantify the effectiveness of the models.

\paragraph{Training Process}
The models were trained using a supervised learning framework with the following setup:
\begin{itemize}
    \item \textbf{Loss Function:} Binary Cross-Entropy (BCE) was employed to optimize the models, defined as:
    \[
    \mathcal{L}_\text{BCE} = - \frac{1}{N} \sum_{i=1}^{N} \left[ y_i \log(\hat{y}_i) + (1 - y_i) \log(1 - \hat{y}_i) \right]
    \]
    where \(y_i\) is the ground truth, \(\hat{y}_i\) is the predicted probability, and \(N\) is the total number of pixels.

    \item \textbf{Optimizer:} Adam optimizer was used with a learning rate of 1e-5 to ensure efficient convergence.

    \item \textbf{Batch Size and Epochs:} Models were trained with a batch size of 4 over 150 epochs to balance computational efficiency and performance.

    \item \textbf{Weight Initialization:} Weight normalization was applied to the convolutional layers to accelerate convergence.

    \item \textbf{Data Augmentation:} To improve generalization, the training data was augmented with random rotations, flips, and intensity variations.
\end{itemize}

\paragraph{Evaluation Metrics}
To evaluate the segmentation performance, the Dice Similarity Coefficient (DSC) was used. This metric quantifies the overlap between the predicted segmentation and the ground truth mask:  
\[
DSC = \frac{2 |A \cap B|}{|A| + |B|}
\]  
where \(A\) represents the predicted segmentation and \(B\) denotes the ground truth.





\section{Results}

All models evaluated in this study follow the autoencoder-based encoder-decoder framework, incorporating architectural innovations to address the unique challenges of lung nodule segmentation. A detailed comparison of the results is presented in Table~\ref{tab:dice-score}, highlighting the strengths and weaknesses of each architecture.

\begin{table}[h]
    \centering
    \caption{Performance Comparison of Segmentation Models (Dice Score)}
    \label{tab:dice-score}
    \begin{tabular}{|l|c|}
        \hline
        \textbf{Model} & \textbf{Dice Score} \\
        \hline
        DPLinkNet50 + Channel Attention Bridge & 0.86 \\
        LinkNet & 0.84 \\
        DPLink & 0.78 \\
        U-Net + Channel Attention Bridge & 0.68 \\
        U-Net + VPM + Channel Attention Bridge & 0.66 \\
        ResDense with Channel Attention & 0.65 \\
        Dilated U-Net & 0.5773 \\
        \hline
    \end{tabular}
\end{table}

\begin{figure}[H]w
    \centering
    \includegraphics[width=0.5\textwidth]{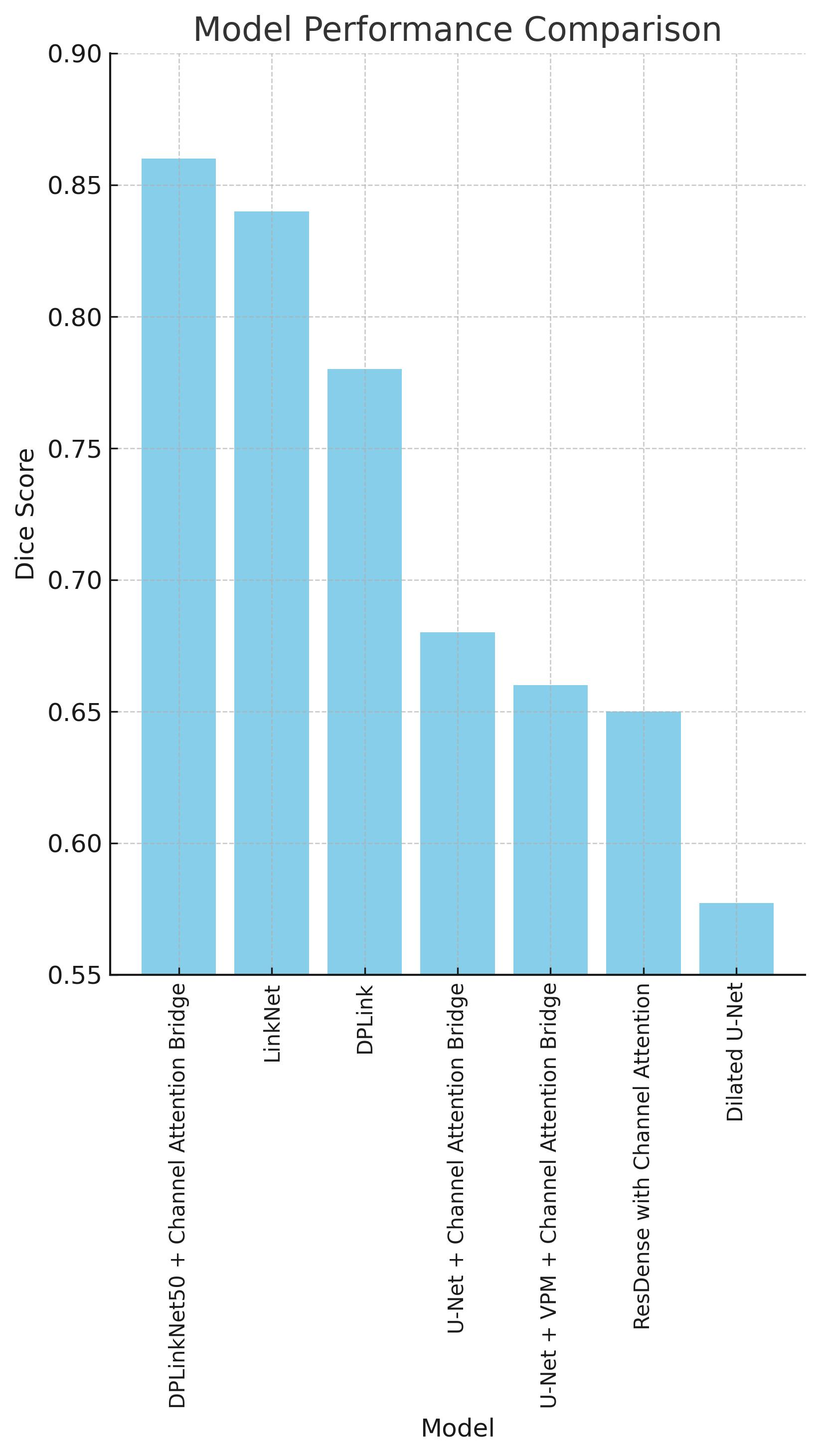} 
    \caption{Performance Comparison of Segmentation Models (Dice Score)}
    \label{fig:dice-bar}
\end{figure}

As shown in Table~\ref{tab:dice-score} and Figure~\ref{fig:dice-bar}, the performance comparison of various segmentation models highlights the differences in their ability to segment lung nodules, measured using the Dice score. The DPLinkNet50 + Channel Attention Bridge model outperforms all others with a Dice score of 0.86, as illustrated in both the table and figure. This model demonstrates the highest accuracy in capturing and segmenting lung nodules, showcasing the effectiveness of the proposed architectural innovations.

\section{Discussion}

This study demonstrates the effectiveness of advanced deep learning architectures and robust preprocessing for lung nodule segmentation. The DPLinkNet50 with Channel Attention Bridge achieved the highest performance, with a Dice score of 0.86, emphasizing the value of attention mechanisms in improving segmentation accuracy. Preprocessing techniques, including K-means clustering and anisotropic diffusion, were critical in enhancing input quality, allowing high performance even on a filtered subset of 356 samples from the LIDC-IDRI dataset.

The proposed models, including U-Net variants and DPLinkNet50, highlight the benefits of architectural innovations such as attention mechanisms and advanced decoders. However, the reliance on a small dataset and the computational demands of certain models pose challenges to broader applicability. Further validation on external datasets is necessary to ensure robustness and generalizability.

To promote transparency and reproducibility, the code for all implemented models, preprocessing steps, and evaluation protocols has been made publicly available on GitHub at  \url{https://github.com/nimashafiei/UNet-Segmentation-Framework}. This enables other researchers to reproduce the results, adapt the pipeline for different datasets, and further explore improvements.

This work underscores the potential of combining data efficiency with architectural innovations to advance computer-aided diagnosis. Future directions include enhancing generalizability through domain adaptation, optimizing models for real-time clinical deployment, and incorporating interpretability to improve their practical utility.

\end{document}